\documentclass[10pt,twocolumn,letterpaper]{article}

\usepackage{wacv}              

\usepackage{graphicx}
\usepackage{amsmath}
\usepackage{amssymb}
\usepackage{booktabs}
\makeatletter
\@namedef{ver@everyshi.sty}{}
\makeatother
\usepackage{tikz}

\usepackage[pagebackref,breaklinks,colorlinks]{hyperref}

\usepackage[capitalize]{cleveref}
\crefname{section}{Sec.}{Secs.}
\Crefname{section}{Section}{Sections}
\Crefname{table}{Table}{Tables}
\crefname{table}{Tab.}{Tabs.}

\usepackage{algorithm,algpseudocode}

\usepackage{multirow}

\usepackage{xpatch}
\usepackage{colortbl}
\usepackage{array,boldline,makecell}
\usepackage{arydshln}
\usepackage{pgfplots}
\usepackage{sidecap, caption}
\usepackage{stackengine}
\usepackage{xspace}
\newcommand\bhr[1]{\textcolor [rgb]{1,0,0}{#1}}
\newcommand\bhb[1]{\textcolor [rgb]{0,0,1}{#1}}

\DeclareMathOperator{\sign}{sign}

\def\wacvPaperID{1339} 
\def\confName{WACV}
\def\confYear{2024}

\begin{document}

\title{Randomized Adversarial Style Perturbations for Domain Generalization}

\author{$\text{Taehoon Kim}^{1}$ \quad\quad\quad \;$\text{Bohyung Han}^{1,2}$\\
$\text{ECE}^{1}$ \& $\text{IPAI}^{2}$, Seoul National University\\
{\tt\small \{kthone, bhhan\}@snu.ac.kr}
}
\maketitle


\begin{abstract}
We propose a novel domain generalization technique, referred to as Randomized Adversarial Style Perturbation (RASP), which is motivated by the observation that the characteristics of each domain are captured by the feature statistics corresponding to its style.
The proposed algorithm perturbs the style of a feature in an adversarial direction towards a randomly selected class.
By incorporating the perturbed styles into training, we prevent the model from being misled by the unexpected styles observed in unseen target domains.
While RASP is effective for handling domain shifts, its na\"ive integration into the training procedure is prone to degrade the capability of learning knowledge from source domains due to the feature distortions caused by style perturbation.
This challenge is alleviated by Normalized Feature Mixup (NFM) during training, which facilitates learning the original features while achieving robustness to perturbed representations.
We evaluate the proposed algorithm via extensive experiments on various benchmarks and show that our approach improves domain generalization performance, especially in large-scale benchmarks.
\end{abstract}



\section{Introduction}
\label{sec:intro}

One of the major drawbacks of machine learning models compared to human intelligence is the lack of adaptivity to distribution shifts.
While humans easily make correct decisions even on unseen domains, deep neural networks often exhibit significant performance degradation on the data from unseen domains.
The lack of robustness to novel domains restricts the applicability of neural networks to real-world problems since it is implausible to build a training dataset that covers all possible domains and follows the true data distribution.
Therefore, learning domain-invariant representations with limited data in the source domains is critical for deploying deep neural networks in practical systems.

Domain Generalization (DG) attempts to train a machine learning model that is robust to unseen target domains, using data from source domains. 
The most straightforward way to achieve this goal is to expose the model to various domains during the training procedure.
To stretch the coverage of the source domains, recent approaches often employ data generation strategies~\cite{shankar2018generalizing, volpi2018generalizing, zhou2020learning, zhou2020deep, zhou2021domain,  li2021simple, nuriel2021permuted, yang2021adversarial, zhong2022adversarial}.
While they have shown promising results on generalization ability, many of them require additional information about data such as domain labels for individual instances~\cite{shankar2018generalizing, zhou2020learning, zhou2020deep, zhou2021domain} or even extra network components such as generators and domain classifiers~\cite{shankar2018generalizing, zhou2020learning, zhou2020deep, yang2021adversarial}. 
However, the additional information including domain labels is unavailable in general and the need for architectural support increases computational complexity and training burden.
There exist a few approaches that do not require extra information about data or additional network modules~\cite{li2021simple, nuriel2021permuted}, but they are limited to straightforward feature augmentations by stochastically adding trivial noise.
While~\cite{shankar2018generalizing, volpi2018generalizing} adopt adversarial data augmentation techniques, they impose perturbations in image space without style disentanglement, which incurs higher computational complexity and inferior generalization capabilities.

This paper presents a simple yet effective data augmentation technique based on adversarial attacks in the feature space for domain generalization.
The proposed approach does not require architectural modifications or domain labels but relies on feature statistics in the intermediate layers.
Our work is motivated by the observation that each visual domain differs in its feature statistics given by instance normalization, which corresponds to the style of a feature.
Based on this observation, existing works attempt to learn style-agnostic networks robust to domain shifts via style augmentations~\cite{zhou2021domain, nuriel2021permuted, kang2022style}.
Although they do not require additional networks~\cite{zhou2021domain, nuriel2021permuted}, they are limited to using simple augmentation techniques with no feedback loop in the augmentation process, leading to suboptimal performance.
While StyleNeophile~\cite{kang2022style} augments novel styles that have different distributions from the source domain using the information observed in the previous iterations, its simple style diversification objective is weak for improving performance on unseen styles in the target domain.

Although our approach follows the same assumption as~\cite{zhou2021domain, nuriel2021permuted, kang2022style}, its objective for style augmentation is unique in the sense that it actively synthesizes hard examples to improve trained models.
To this end, inspired by adversarial attacks~\cite{goodfellow2014explaining, madry2018towards,xie2019improving}, the proposed method, referred to as Randomized Adversarial Style Perturbations (RASP), adversarially augments the styles of features so that the corresponding examples deceive the network to be misclassified.
Unlike the other methods based on adversarial attacks toward the fixed target label~\cite{shankar2018generalizing, zhong2022adversarial}, we draw random labels for attacks to ensure the plausibilty of the augmented styles. 
While the features with modified styles strengthen the generalization ability on unseen domains, they might neglect crucial information observable in the source domains since the style augmentation is prone to disturb the representations of perturbation-free examples.
To compensate for this, we propose Normalized Feature Mixup (NFM) technique based on mixup~\cite{zhang2018mixup}. 
Instead of applying the na\"ive feature mixup technique, NFM combines the normalized representations of perturbed and perturbation-free features.
By integrating normalized features given by NFM, we successfully maintain the representations from the source domains while taking advantage of style augmentation based on RASP.

Our contributions are summarized as follows:
\begin{itemize}
\item We present a unique style augmentation technique, referred to as RASP, for domain generalization based on adversarial learning.
This method is free from any architectural modifications or the need for the domain label of each example. 
\item We introduce a novel feature mixup method, NFM, which allows us to maintain knowledge from source domains while facilitating the adaptation to fresh data via robust domain augmentation.
\item The proposed approach consistently demonstrates outstanding generalization ability in multiple standard benchmarks, especially in large-scale datasets.
\end{itemize}

The rest of this paper is organized as follows.
We first review previous works about domain generalization in Section~\ref{sec:related}, and our main algorithm based on RASP and NFM is discussed in Section~\ref{sec:method}.
We present experimental results from the standard benchmarks in Section~\ref{sec:experiments} and conclude this paper in Section~\ref{sec:conclusion}.


\section{Related Works}
\label{sec:related}
\begin{figure*}[t]
  \centering
   \includegraphics[width=\linewidth]{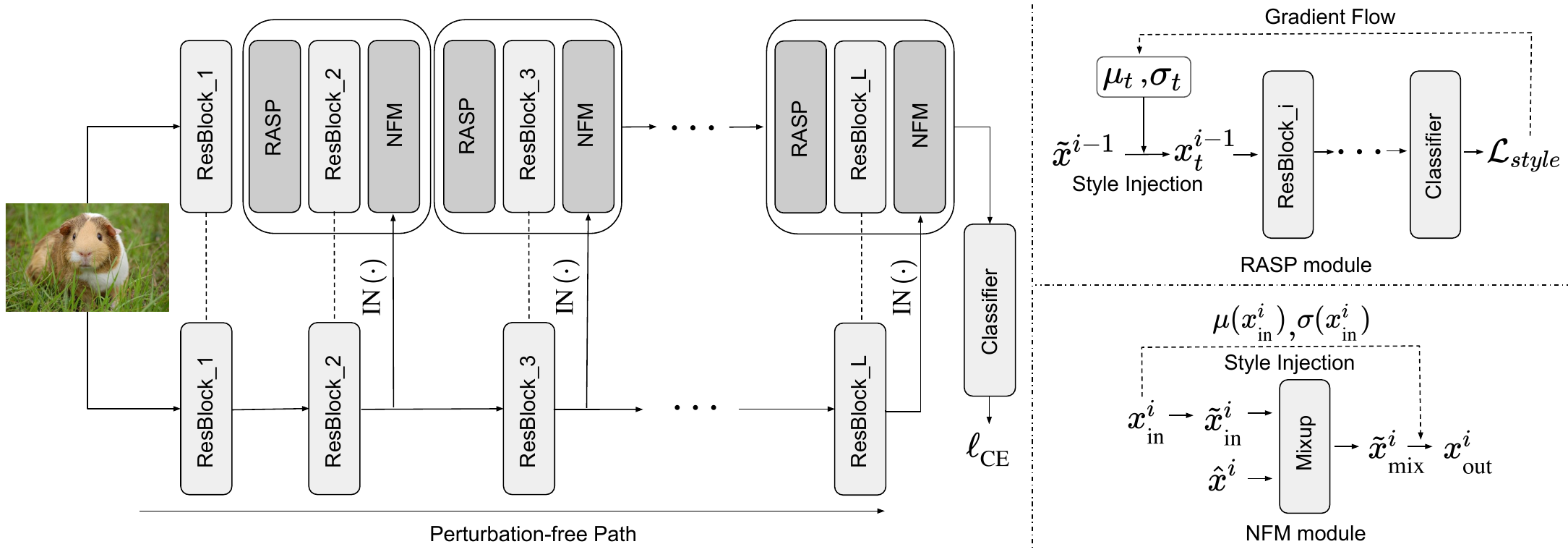}
   \caption{Overall framework of our algorithm, Randomized Adversarial Style Perturbations (RASP) with Normalized Feature Mixup (NFM). Our model runs two parallel paths, one with RASP+NFM and the other without the modules.
   Before each block in the RASP+NFM path, we apply the RASP module to augment novel styles. RASP adjusts the style of each feature by minimizing the loss with respect to a random target class different from the ground-truth. After passing through the RASP module, followed by a ResBlock, NFM is performed to regularize the style-augmented features. The perturbation-free path injects the style normalized information into the NFM module. Note that the common network components in the two paths share their parameters.}
   \label{fig:overall}
\end{figure*}

In the pursuit of robust Domain Generalization (DG), a central challenge is to develop models that generalize to unseen target domains using only source domains for training.
Existing approaches address the DG problem by using the following techniques: 1) meta-learning 2) data augmentation, 3) feature statistics manipulation, and 4) flat minima seeking.
This section summarizes the technical details of each of the four categories.

\vspace{-2mm}
\paragraph{Meta-learning approaches}
The algorithms in this line of research formulate the domain generalization task as a meta-learning problem~\cite{li2018learning, balaji2018metareg, du2020learning} by splitting the source domains into the meta-train and meta-test sets. Using these sets, they adopt the learn-to-learn schemes of meta-learning for the generalization on unseen domains.
However, they rely heavily on the assumption that the diversity of the source domains is large enough to effectively cover unseen domains, which may not hold in real-world scenarios.

\vspace{-2mm}
\paragraph{Data augmentation approaches}
The methods in this category, which deal with domain shifts by introducing new images belonging to new domains for training~\cite{zhou2020learning, zhou2020deep, yang2021adversarial, xu2021fourier, shankar2018generalizing, tarvainen2017mean, volpi2018generalizing, zhong2022adversarial}, and can be divided into two groups.
One group uses generative models to increase the number of training examples~\cite{zhou2020learning, zhou2020deep, yang2021adversarial}, which induces extra computational complexity and instability of training.
The approaches in the other direction rely on image perturbation for data augmentation~\cite{xu2021fourier, shankar2018generalizing, volpi2018generalizing, zhong2022adversarial}.
However, some of them~\cite{xu2021fourier, shankar2018generalizing} still require auxiliary neural networks for the regularization. 
While some image perturbation approaches~\cite{volpi2018generalizing, shankar2018generalizing, zhong2022adversarial} employ adversarial augmentation techniques similar to our method, they operate on image space without style disentanglement, leading to high computational complexity and inferior performance compared to the proposed method.

\vspace{-2mm}
\paragraph{Feature statistics manipulation approaches}
The methods in this type take advantage of the observation that the feature statistics capture characteristics of the visual domains~\cite{zhou2021domain, nuriel2021permuted, kang2022style, seo2020learning}.
DSON~\cite{seo2020learning} introduces domain specific normalization layer using the weighted sum of batch and instance normalization statistics.
MixStyle~\cite{zhou2021domain} generates novel domain features by mixing feature statistics from different images while pAdaIN~\cite{nuriel2021permuted} randomly permutes the statistics of each feature before every batch normalization layer. 
Similarly, SFA~\cite{li2021simple} adopts a feature-level augmentation technique that applies simple stochastic linear transforms.
Unlike~\cite{zhou2021domain, nuriel2021permuted}, StyleNeophile~\cite{kang2022style} augments styles that have different distributions from those in previous iterations. 
Although StyleNeophile~\cite{kang2022style} does not rely on simple stochasticity, there is no guarantee that the augmented styles will be useful for the generalization in target tasks. 

\vspace{-2mm}
\paragraph{Flat minima seeking approach}
It is well-known that flat minima in the objective function facilitates learning robust models to the variations of input data.
SWAD~\cite{cha2021swad} exploits this property and proposes the stochastic weight averaging strategy~\cite{izmailov2018averaging} tailored for domain generalization; the technique finds flat minima for enhancing the generalization ability to domain shift.

\section{Proposed Approach}
\label{sec:method}
This section describes the technical details of the proposed Randomized Adversarial Style Perturbations (RASP) and the Normalized Feature Mixup (NFM) methods.

\subsection{Background}
\paragraph{Instance normalization and style}
Recent studies on neural style transfer~\cite{huang2017arbitrary, ulyanov2016instance} discover that the style of a feature can be captured by the instance-specific channel-wise mean and standard deviation.
Instance Normalization (IN)~\cite{ulyanov2016instance} removes the effect of styles on features by normalizing features as follows:
\begin{equation}
\text{IN}(z) = \gamma \dfrac{z-\mu(z)}{\sigma(z)} + \beta,
\end{equation}
where $z$ is the feature for an input example $x$, $(\gamma, \beta)$ are learnable affine parameters, and $(\mu(z), \sigma(z))$ are instance-specific channel-wise statistics computed by
\begin{align}
\mu(z) &= \dfrac{1}{HW}\sum_{h=1}^{H}\sum_{w=1}^{W}z_{h,w}, \\
\sigma(z) &= \sqrt{\dfrac{1}{HW}\sum_{h=1}^{H}\sum_{w=1}^{W}(z_{h,w}-\mu(z))^2}.
\end{align}
AdaIN~\cite{huang2017arbitrary} allows the style of an input feature $z'$ to be transferred to the content of another input $z$ by replacing the affine parameters of IN with the feature statistics of the style image $z'$ as follows:
\begin{equation}
\text{AdaIN}(z, z') = \sigma(z') \dfrac{z-\mu(z)}{\sigma(z)} + \mu(z').
\end{equation}

\paragraph{Gradient-based attacks}
Adversarial attacks~\cite{kurakin2016adversarial, zhao2018generating,moosavi2016deepfool, xie2019improving, madry2018towards, goodfellow2014explaining} trick a neural network by injecting imperceptible small noise to an example.
A popular way to generate such perturbations is to utilize the gradient information of the network.
Given an image-label pair $(x,y)$ and model parameters $\theta$, FGSM~\cite{goodfellow2014explaining} computes the optimal perturbation of a linearized cost function, which is given by
\begin{equation}
x = x + \epsilon \sign(\nabla_{x}J_\theta(x,y)),
\end{equation}
where $\epsilon$ is the magnitude of perturbation and $J_\theta(\cdot, \cdot)$ is a task-specific loss function.
I-FGSM~\cite{kurakin2016adversarial} extends FGSM~\cite{goodfellow2014explaining} by using multiple iterations as
\begin{align}
x_{0} &= x, \\ 
x_{t+1} & = \text{clip}_{x, \epsilon}\left[ x_t + \alpha\sign(\nabla_{x} J_\theta(x_t, y)) \right],
\end{align}
where $\alpha$ is the step size for each iteration and $\text{clip}_{x, \epsilon}[\cdot]$ denotes a clipping operation that restricts the magnitude of a perturbation from the original image only up to $\epsilon$.
We adopt an unclipped and targeted version of I-FGSM as our baseline of adversarial attacks.

\subsection{Overall Framework}
Let $(x, y)$ denote an image and class label pair sampled from an arbitrary source domain. 
We adopt Empirical Risk Minimization (ERM) using the standard cross entropy loss, $\ell_\text{CE}$, as our baseline.
Our goal is to train a feature extractor $f_{\theta}$ and a classifier $g_{\phi}$ parameterized by $\theta$ and $\phi$, respectively, which are robust to domain shifts. 
To apply our methods, we divide $f_{\theta}$ into $L$ residual blocks as $  f_{\theta}:= f_{\theta}^{L} \circ f_{\theta}^{L-1} \circ \cdots \circ f_{\theta}^{1}$.
When we train the proposed network, we consider an additional forwarding path---a perturbation-free path---with shared weights in parallel and make the two paths interact with each other.

To introduce the challenging styles during training, we incorporate the Randomized Adversarial Style Perturbations (RASP) module, which will be discussed in Section~\ref{ssec:asa}, before each block of the backbone network with a probability of 0.5.
When RASP is applied, we also perform Normalized Feature Mixup (NFM) after each block with a probability of 0.5.
It prevent the augmented features from being deviated too much from the original features.
Figure~\ref{fig:overall} illustrates the overall framework of the proposed approach.

\begin{table*}[t]
\caption{Performance on DomainNet with two different backbone networks. RASP+NFM presents outstanding accuracy in this large-scale dataset. The bold-faced numbers indicate the best performance.}
\label{tab:domain}
    \vspace{-5mm}
    \begin{center}
        \scalebox{0.9}{
	\setlength\tabcolsep{5pt}
	\hspace{-3mm}
	    \begin{tabular}{ccccccccc}
            \multicolumn{9}{c}{(a) Results of ResNet18 on DomainNet} \\  \toprule
            \multicolumn{1}{c}{Method}& \multicolumn{1}{c}{Additional components}& \multicolumn{1}{c}{Clipart} & \multicolumn{1}{c}{Infograph} & \multicolumn{1}{c}{Painting} & \multicolumn{1}{c}{Quickdraw} & \multicolumn{1}{c}{Real} &  \multicolumn{1}{c}{Sketch} & \multicolumn{1}{c}{Avg.}  \\ \midrule
            ERM &---   & 56.6 & 18.4 & 45.3 & 12.5 & 57.9 & 38.8 & 38.3 \\ 
            MetaReg~\cite{balaji2018metareg} &Domain label, Task network& 53.7 & 21.1 & 45.3 & 10.6 & \textbf{58.5} & 42.3 & 38.6 \\ 
            DMG~\cite{chattopadhyay2020learning}& Domain label, Mask predictor& 60.1 & 18.8 & 44.5 & 14.2 & 54.7 & 41.7 & 39.0 \\ 
            StyleNeophile~\cite{kang2022style}&---   & 60.1 & 17.8 & 46.5 & 14.6 & 55.4 & 45.3 & 40.0 \\
            RASP+NFM (ours) &---     &\textbf{60.4 $\pm$ 0.2} & \textbf{22.6$\pm$ 0.1}   & \textbf{50.2$\pm$ 0.2} & \textbf{17.2$\pm$ 0.3} & 56.8$\pm$ 0.3 &  \textbf{48.5$\pm$ 0.4} & \textbf{42.6} \\ \toprule
            \multicolumn{9}{c}{} \\

            \multicolumn{9}{c}{(b) Results of ResNet50 on DomainNet} \\  \toprule
            \multicolumn{1}{c}{Method} & \multicolumn{1}{c}{Additional components}& \multicolumn{1}{c}{Clipart} & \multicolumn{1}{c}{Infograph} & \multicolumn{1}{c}{Painting} & \multicolumn{1}{c}{Quickdraw} & \multicolumn{1}{c}{Real} & \multicolumn{1}{c}{Sketch} & \multicolumn{1}{c}{Avg.} \\ \midrule
            ERM &---   &64.0 & 23.6 & 51.0 & 13.1 & 64.5 & 47.8 & 44.0 \\
            MetaReg~\cite{balaji2018metareg}&Domain label, Task network & 59.8 & 25.6 & 50.2 & 11.5 & 65.5 & 50.1 & 43.6 \\ 
            DMG~\cite{chattopadhyay2020learning}&Domain label, Mask predictor & 65.2  & 22.2 & 50.0 &15.7 & 59.6 & 49.0 & 43.6 \\ 
            StyleNeophile~\cite{kang2022style}&---    & 66.1 & 21.4 & 51.4 & 15.3 & 61.7 & 51.8 & 44.6 \\
            SWAD~\cite{cha2021swad}& ---   & 66.0 & 22.4 & 53.5 & 16.1 & \textbf{65.8} & 55.5 & 46.5 \\
            RASP+NFM (ours)& ---    &\textbf{66.5$\pm$ 0.2} & \textbf{27.4$\pm$ 0.2}  & \textbf{55.2$\pm$ 0.2} & \textbf{16.9$\pm$ 0.3} & 63.7$\pm$ 0.1 &  \textbf{53.8$\pm$ 0.3} & \textbf{47.2}  \\ \bottomrule				 
	    \end{tabular} 
	}
	\vspace{-2mm}
    \end{center}
\end{table*}

%

\subsection{Randomized Adversarial Style Perturbations (RASP)}
\label{ssec:asa}
The proposed method augments the styles of features in each block of a network, employs the style-augmented examples for adversarial training, and improves the robustness of the trained model to the features regardless of their styles.
While the existing style perturbation methods~\cite{zhou2021domain, nuriel2021permuted, kang2022style} are helpful for learning style-agnostic feature extractors, we argue that simply generating diverse examples is suboptimal for training models and that the target direction in style augmentation is critical for training domain-agnostic models using a limited number of data with new styles.

RASP generates examples that meet two desirable properties---style difficulty and plausibility.
First, the augmentation process is conducted in a way that provides challenging styles, to prevent the model trained with the styles from being deceived by the unexpected distribution shifts.
Second, since the proposed algorithm perturbs an example towards one of the target classes other than its ground-truth, the generated examples tend to be more realistic than those obtained by simply reducing the score corresponding to the ground-truth label. 
In addition to the target selection strategy, we adopt a threshold to terminate the RASP iterations when the prediction score for the ground-truth label falls below the threshold; it ensures that the augmented styles are within the desired range aligned with realistic images.

The training procedure with RASP is simple.
Given a feature $z^{i-1}$ for an input $x$ after the $(i-1)^{\text{st}}$ block and a randomly sampled target label $y_\text{target}$ different from the original label $y$, we compute the style loss $\mathcal{L}_\text{style}$ at each attack iteration $t$ as
\begin{align}
\mathcal{L}_\text{style} = \ell_\text{CE} \left( g_{\phi} \circ f_{\theta}^{L} \circ \cdots \circ f_{\theta}^{i}(z_t^{i-1}), y_\text{target} \right),
\end{align}
where $z_t^{i-1} = \tilde{z}^{i-1} \cdot \sigma_{t} + \mu_{t}$ is the denormalized vector and $\ell_\text{CE}$ is standard crossentropy loss.
Notice that $\tilde{z}^{i-1}$ is the instance normalized feature derived from $z^{i-1}$ and $(\mu_t, \sigma_t)$ characterizes the style at the $t^\text{th}$ iteration, where $(\mu_0, \sigma_0) = (\mu(z^{i-1}), \sigma(z^{i-1}))$.
Given a step size $\epsilon$, the style of a feature is updated in a way that decreases $\mathcal{L}_\text{style}$ as follows:
\begin{align}
\mu_{t+1} &= \mu_{t} - \epsilon \cdot \lVert \mu_0 \rVert_{2} \cdot \sign( \nabla_{\mu_t} \mathcal{L}_\text{style}) 
\label{eq:updates_mean} \\
 \sigma_{t+1} &= \sigma_{t} - \epsilon \cdot \lVert \sigma_0 \rVert_{2} \cdot \sign( \nabla_{\sigma_t} \mathcal{L}_\text{style})
\label{eq:updates_sigma}
\end{align}
as long as the following condition is met:
\begin{equation}
\text{softmax}(g_{\phi} \circ f_{\theta}^{L} \circ \cdots \circ f_{\theta}^{i}(z_t^{i-1}))_{y} \geq \tau.
\end{equation}
where $\tau$ is the score threshold of the ground-truth class for terminating the RASP iterations.
Note that the step sizes in RASP for updating the mean and the standard deviation are proportional to their magnitudes as in \eqref{eq:updates_mean} and \eqref{eq:updates_sigma}.

\subsection{Normalized Feature Mixup (NFM)}
Although RASP provides plenty of effective novel styles that strengthen generalization performance on unseen domains, it degrades the capability of learning features from source domains since perturbed styles may deviate excessively from the original ones.
To compensate for this phenomenon, we propose the Normalized Feature Mixup technique (NFM) technique, which ensembles instance-normalized features from both the perturbation-free path and the RASP path.
By doing this, NFM preserves the knowledge from the source domain while learning robust representations by taking advantage of style augmentations.

Instead of using the features before the instance normalization, which may change the augmented styles back to the original one by mixup, NFM performs mixup with the normalized features (content features) and then applies augmented styles to the mixed normalized features via denormalization. 
Formally, given a feature from the $i^{\text{th}}$ block $z^{i}_{\text{in}}$, we obtain an instance-normalized feature, $\tilde{z}^{i}_{\text{in}}$, and its augmented style obtained from the RASP path.
NFM module mixes $\tilde{z}^{i}_{\text{in}}$ with an instance-normalized feature from the perturbation-free path in the $i^{\text{th}}$ block, $\hat{z}^{i}$, to get a mixed normalized feature, $\tilde{z}^{i}_{\text{mix}}$, and denormalizes it with the augmented style $(\mu(z^{i}_{\text{in}}), \sigma(z^{i}_{\text{in}}))$  to obtain the final output, $z^i_{\text{out}}$, as follows:
\begin{align}
\tilde{z}^i_{\text{mix}} &= \alpha\cdot \hat{z}^i + (1-\alpha)\cdot \tilde{z}^i_{\text{in}}, \\
z^{i}_{\text{out}} &= \tilde{z}^i_{\text{mix}} \cdot \sigma(z^{i}_{\text{in}}) + \mu(z^{i}_{\text{in}}),
\end{align}
where $\alpha \sim \text{Beta}(0.1,0.1)$ determines the mixup ratio.

\subsection{Inference}
While our method utilizes an additional forwarding path and optimization steps during training, we only use the perturbation-free path for inference.
Therefore, it does not incur additional computational cost at the inference time.

\section{Experiments}
\label{sec:experiments}

We demonstrate the performance of the proposed algorithm on standard benchmarks of domain generalization and analyze the characteristics of our approach in comparison with existing techniques.

\subsection{Datasets and Evaluation Protocol}
We evaluate the proposed algorithm on DomainNet~\cite{peng2019moment}, Office-Home~\cite{venkateswara2017deep}, and PACS~\cite{li2017deeper}, which are standard benchmarks for domain generalization.
We set DomainNet, which is a large-scale dataset in terms of the number of classes and examples, as our primary target benchmark since verification in a large-scale benchmark is essential to confirm whether the proposed algorithm is applicable in real-world problems. 
DomainNet contains 586,475 images of 6 domains (Clipart, Infograph, Painting, Quickdraw, Real, and Sketch) and 345 classes.
Office-Home contains 15,558 images of 65 classes from 4 different domains (Artistic, Clipart, Product, and Real world) while PACS, the smallest dataset, consists of 9,991 images of 4 domains (Photo, Art paint, Cartoon, and Sketches) and 7 classes (dog, elephant, giraffe, guitar, horse, house and person).
\begin{table*}[t]
\caption{Performance on Office-Home with two different backbone networks. Note that the shaded rows indicate the algorithms that use different hyperparameters for individual target domains, resulting in overestimated accuracies. The bold-faced numbers indicate the best performance among the methods under fair comparisons without domain-specific hyperparameter turning.}
\vspace{-5mm}
\label{tab:off}
    \begin{center}
        \scalebox{0.9}{
	\setlength\tabcolsep{12pt}
	
	\hspace{-3mm}
	    \begin{tabular}{ccccccc}
            \multicolumn{7}{c}{(a) Results of ResNet18 on Office-Home} \\  \toprule
             \multicolumn{1}{c}{Method} &\multicolumn{1}{c}{Additional components}& \multicolumn{1}{c}{Art} & \multicolumn{1}{c}{Clipart} & \multicolumn{1}{c}{Product} & \multicolumn{1}{c}{Real} & \multicolumn{1}{c}{Avg.} \\ \midrule
            ERM &---   & 59.0 & 48.4 & 72.5 & 75.5 & 63.9 \\
            CrossGrad~\cite{shankar2018generalizing}&Domain classifier/label & 58.4 & 49.4 & 73.9 & 75.8 & 64.4 \\
            MixStyle~\cite{zhou2021domain}&Domain label & 58.7 & 53.4 & 74.2 & 75.9 & 65.5 \\
            StyleNeophile~\cite{kang2022style}&---    & 59.6 & 55.0 & 73.6 & 75.5 & 65.9  \\
            RASP+NFM (ours) & ---    &\textbf{59.7 $\pm$ 0.4}& \textbf{57.6$\pm$ 0.9} & \textbf{75.2$\pm$ 0.3} & \textbf{76.7$\pm$ 0.4} & \textbf{67.3} \\\midrule
            \rowcolor{gray!10} DDAIG~\cite{zhou2020deep}&Generator & 59.2  & 52.3 & 74.6 & 76.0 & 65.5 \\
            \rowcolor{gray!10} ADVTSRL~\cite{yang2021adversarial} &Generator& 60.7 & 52.9 & 75.8 & 77.2 & 66.7 \\ \toprule 
           \multicolumn{7}{c}{} \\

            \multicolumn{7}{c}{(b) Results of ResNet50 on Office-Home} \\ \toprule
            \multicolumn{1}{c}{Method} &\multicolumn{1}{c}{Additional components}& \multicolumn{1}{c}{Art} & \multicolumn{1}{c}{Clipart} & \multicolumn{1}{c}{Product} & \multicolumn{1}{c}{Real} & \multicolumn{1}{c}{Avg.} \\ \midrule
            ERM &---   & 64.7 & 58.8 & 77.9 & 79.0 & 70.1 \\ 
            CrossGrad~\cite{shankar2018generalizing} &Domain classifier/label& 67.7 & 57.7 & 79.1 & 80.4 & 71.2 \\
            MixStyle~\cite{zhou2021domain} &Domain label& 64.9 & 58.8 & 78.3 & 78.7 & 70.2 \\
            SWAD~\cite{cha2021swad} &---   &66.1&57.7&78.4&80.1&70.6 \\
            RASP+NFM (ours)&---    & \textbf{68.8$\pm$ 0.4} & \textbf{61.7$\pm$ 1.0} & \textbf{79.8$\pm$ 0.2} & \textbf{80.9$\pm$ 0.3} & \textbf{72.8} \\ \midrule
            \rowcolor{gray!10} DDAIG~\cite{zhou2020deep}&Generator & 65.2 & 59.2 & 77.7 & 76.7 & 69.7 \\
            \rowcolor{gray!10} ADVTSRL~\cite{yang2021adversarial}&Generator & 69.3 & 60.1 & 81.5 & 82.1 & 73.3 \\ \bottomrule
            \end{tabular} 
        }
        
    \end{center}
    
\end{table*}

We use the source domain validation set as the validation set for model selection, following the \textit{``training-domain validation''} criterion of DomainBed~\cite{gulrajani2021search}.
All the results are average classification accuracy over five runs with different random seeds.

\subsection{Implementation Details}
\label{sec:imp}
We employ ResNet18 or ResNet50~\cite{he2016deep} pretrained on ImageNet~\cite{deng2009imagenet} as our backbone network architectures. 
We use the SGD optimizer with a learning rate of 0.0005 decayed by 0.1 after 30 epochs.
The batch size is set to 32 and the number of epochs for training is set to 60.

For our approach, we set the threshold of the ground-truth class probability $\tau$ to 0.8, the step size $\epsilon$ to $\frac{2}{255}$, and the number of attack iterations to 5 for all datasets.
Note that $\epsilon$ is rescaled by multiplying $\frac{64}{\text{channel size}}$ to maintain the size of perturbations at each layer.
RASP and NFM are applied to the 2nd, 3rd, and 4th residual blocks, before and after each block, respectively, as illustrated in Figure~\ref{fig:overall}.
All of the experiments are performed on a single TITAN-XP GPU using VESSL~\cite{vessl}.

\subsection{Results on DomainNet}
We present the quantitative results of RASP with NFM on DomainNet in comparison with other existing methods. 
DomainNet is the most challenging dataset for DG since it is the largest and there are substantial domain shifts between domains, compared to other datasets.
Despite these challenges, the proposed algorithm consistently achieves significant performance gains in all cases except the Real domain with the ResNet18 and ResNet50 backbones as shown in Table~\ref{tab:domain}.
Note that improvements in the domains with severe distribution shifts (Infograph, Painting, Quickdraw, Sketch) are more salient than the mild domains (Clipart, Real).
This result implies that the optimization for the worst-case styles is particularly helpful in the case that  there is a large domain gap between source and target domains.

 
 \begin{table*}[t]
	\caption{Performance on PACS with the ResNet18 backbone network.
	Note that the shaded rows of the table indicate the algorithms that use different hyperparameters for individual target domains. The bold-faced numbers indicate the best performance among the methods that do not require the hyperparameter tuning specific to target domains.}
	\vspace{-5mm}
	\label{tab:pacs}
	
	\begin{center}
		\scalebox{0.9}{
			\setlength\tabcolsep{12pt}
			\begin{tabular}{ccccccc}
			   \multicolumn{7}{c}{Results of ResNet18 on PACS} \\  \toprule			 
			Method &Additional components& Art & Cartoon  &Photo &Sketch&   Avg.  \\ \midrule 
	 			ERM            & ---                                                    & 75.1   &  74.2  & 95.6 & 68.4 & 78.3 \\ 
				CrossGrad~\cite{shankar2018generalizing}   &Domain classifier/label        & 79.8.  & 76.8 & 96.0 & 70.2 & 80.7 \\ 
				MixStyle~\cite{zhou2021domain}         &Domain label                  & 84.1  &78.8   & \textbf{96.1} &75.9& 83.7\\
				StyleNeophile~\cite{kang2022style}    &---                   & 84.4  &78.4   & 94.9 & \textbf{83.3} &\textbf{85.5}  \\
				RASP+NFM (ours)                              &---                        & \textbf{84.6$\pm$ 0.5}  &\textbf{79.8$\pm$ 0.5}   & 94.1$\pm$ 0.4 & 80.1$\pm$ 1.1&84.7  \\ \midrule
    \rowcolor{gray!10}DDAIG~\cite{zhou2020deep}                    &Generator              & 84.2  & 78.1 &95.3 &74.7&  83.1 \\ 
    \rowcolor{gray!10}ADVTSRL~\cite{yang2021adversarial}    &Generator               & 85.8  &80.7   & 97.3 & 77.3&85.3  \\
			 				 \bottomrule				 
			\end{tabular} 
		}
	\end{center}
	\vspace{-2mm}
\end{table*}

\subsection{Results on Office-Home}
To validate the effectiveness of our approach, we conducted experiments on an additional dataset, Office-Home.
Table~\ref{tab:off} clearly shows that RASP with NFM outperforms other methods on this dataset.
The proposed method exhibits greater performance improvement in more challenging target domains; this is consistent with the observation in DomainNet that RASP is useful especially when there exists a larger domain discrepancy between the source and the target.  
ADVTSRL~\cite{yang2021adversarial}  and DDAIG~\cite{zhou2020deep} are optimized with a separate set of hyperparameters for each target domain and the comparisons with these methods are unfair.

\begin{table*}[t]
	\caption{Ablation study results on the variations of attack iterations ($T$) and attack termination threshold ($\tau$) with ResNet18 on Office-Home. 
	}
	\vspace{-5mm}
	\label{tab:hyper}
	\begin{center}
	\setlength\tabcolsep{10pt}	
	\scalebox{0.9}{
\begin{tabular}{lccccccc}
\toprule
Ablation types   &  Variations	& Art &Clipart & Product &Real &Avg. &Source Acc.\\ 
\midrule
\multirow{5}{*}{(a) Attack iteration ($T$)}   &1      & 59.5 & 53.0 & 75.6  &76.9 & 66.3 & 83.7  \\
&2       & \textbf{60.8} & 55.0& \textbf{76.0}  & 77.2 & 67.3  & 84.1 \\
&3       & 60.5& 56.5& 75.5  & \textbf{77.4}& \textbf{67.5} & \textbf{84.3}  \\
&4       & 60.2& 56.9& 75.6  & 77.0 & 67.4  & 83.8 \\
&5       & 59.7& \textbf{57.6}& 75.2  & 76.7 & 67.3  & 83.5 \\ \hline
\multirow{5}{*}{(b) Threshold ($\tau$)}           & 0.0           & 57.4& 58.0  & 72.4 & 74.8 &65.7  & 82.4\\
              & 0.2           & 58.0& \textbf{58.6}  & 73.4 & 75.2 &65.7  & 83.2 \\ 
              & 0.4           & 58.6& 58.5  & 74.2 & 76.0 &66.3  & 83.5\\  
               & 0.6           & 59.2& 58.0  & 74.7 & 76.5&67.1  & \textbf{83.8} \\ 
               & 0.8          & \textbf{59.7}& 57.6  & \textbf{75.2} & \textbf{76.7} & \textbf{67.3}  & 83.5 \\ 
\bottomrule
\end{tabular}
}
\end{center}
\vspace{-2mm}
\end{table*}

\begin{table}[t]
	\caption{Ablation study results on the step size ($\epsilon$) with ResNet18 on Office-Home.
	}
	\vspace{-5mm}
	\label{tab:eps}
	\begin{center}
	\setlength\tabcolsep{3pt}	
	\scalebox{0.9}{
\begin{tabular}{ccccccc}
\toprule
        Step size 	& Art &Clipart & Product &Real  &Avg. &Source Acc.\\ 
\midrule
   $\epsilon=0.5/255$      &60.1 & 53.2 & \textbf{76.1}  & 76.9 &66.6  & 83.9  \\
                                   $\epsilon=1/255$                 &\textbf{61.0} & 55.4 & 75.9  & \textbf{77.2} &\textbf{67.4}  & \textbf{84.2} \\
                                     $\epsilon=2/255$               &59.7 & 57.6 & 75.2  & 76.7 &67.3  & 83.5  \\
                                       $\epsilon=3/255$             &59.0 & \textbf{58.2} & 74.0 & 76.0 &66.8  & 83.3  \\ 
\bottomrule
\end{tabular}
}
\end{center}
\vspace{-2mm}
\end{table}

\subsection{Results on PACS}
We also evaluate RASP with NFM on the PACS dataset and present the results in Table~\ref{tab:pacs}.
The proposed method is competitive with all the compared methods even without additional components such as domain classifiers or labels.
As mentioned earlier, ADVTSRL~\cite{yang2021adversarial} and DDAIG~\cite{zhou2020deep} use domain-specific hyperparameters, which makes the comparisons with our method unfair.

\subsection{Ablation studies}
As pointed out in DomainBed~\cite{gulrajani2021search}, selecting proper hyperparameters is a major issue for domain generalization since the target domain data is inaccessible during training.
To ensure the applicability of the proposed algorithm to real-world domain generalization scenarios, we analyze the effect of each hyperparameter on target domain accuracies and source domain validation accuracies. 
Also, we test the various options of the proposed algorithm to further validate the effectiveness of each component.

\vspace{-2mm}
\paragraph{Number of attack iterations}
We study how the performance of RASP is influenced by the number of attack iterations, $T$.
Table~\ref{tab:hyper}(a) presents the evaluation results on Office-Home by varying the number of attack iterations.
Since the augmented examples with more adversarial iterations have more challenging styles, the classification accuracy on relatively easy target domains, {\it e.g.,} Product and Real, decreases as the number of iterations increases while the accuracy on more difficult target domains, \eg, Clipart, benefits from more iterations.
An important observation from Table~\ref{tab:hyper}(a) is that target domain accuracies have positive correlations with source domain validation accuracies; one can easily select the proper $T$ by observing source domain validation accuracies.
\begin{table}[t!]
	\caption{Ablation study results on the location where RASP is applied with ResNet18 on PACS.
	}
	\vspace{-5mm}
	\label{tab:layer}
	\begin{center}
	\setlength\tabcolsep{2pt}	
	\scalebox{0.9}{
\begin{tabular}{ccccccccc}
\toprule
RGB&Res2 & Res3 & Res4 & Art & Cartoon & Photo & Sketch &Avg. \\ 
\midrule
\midrule
&& &    & 75.1   &  74.2  & 95.6 & 68.4 & 78.3  \\ \midrule
ASA~\cite{zhong2022adversarial}& & &   & 75.8 &76.3&95.7&67.4&78.8 \\ \midrule
\shortstack[l]{CrossGrad \\ \quad ~\cite{shankar2018generalizing}}& & &   & 79.8 &76.8&\textbf{96.0}&70.2&80.7 \\ 
\midrule
\checkmark && &   & 79.3 & 76.7 & \textbf{96.0} & 73.5 & 81.4 \\ 
&\checkmark & &   & 82.2 & 77.9 & 95.1 & 76.9 & 83.0 \\ 
&& \checkmark &   & 82.1 & 77.3 & 94.7 & 77.7 & 83.0 \\ 
&& & \checkmark   & 80.1 & 78.0 & 95.5 & 68.2 & 80.5 \\ 
&\checkmark & \checkmark &  & 83.2 & 78.0 & 94.2 & \textbf{80.2} & 83.9 \\ 
&\checkmark & & \checkmark  & 84.5 & 79.2 & 94.7 & 76.4 & 83.7 \\ 
&& \checkmark & \checkmark  & 83.4 & 79.5 & 94.4 & 77.6 & 83.7 \\ 
&\checkmark & \checkmark & \checkmark & \textbf{84.6} & \textbf{79.8} & 94.1  & 80.1 & \textbf{84.7}  \\ 
\bottomrule
\end{tabular}
}
\end{center}
\vspace{-2mm}
\end{table}

\begin{table*}[t!]
	\caption{Ablation study results on other variations of our algorithm with ResNet18 on PACS.
	}
	\vspace{-5mm}
	\label{tab:multi_abl}
	\begin{center}
	\setlength\tabcolsep{10pt}	
	\scalebox{0.9}{
\begin{tabular}{lcccccc}
\toprule
Ablation types   &  Variations	& Art &Cartoon & Photo &Sketch &Avg. \\ 
\midrule
\multirow{2}{*}{(a) Augmentation Objective}   &Ours      & \textbf{84.6} & \textbf{79.8} & \textbf{94.1}  & 80.1 & \textbf{84.7}   \\
&RASP$_\text{GT}$       & 84.3& 78.1& 93.3  & \textbf{80.6} & 84.1   \\ \hline
\multirow{4}{*}{(b) NFM}           & w/o NFM             & 82.6 & 79.3  & 92.9 & \textbf{80.1} &83.7  \\
\multicolumn{1}{c}{}              & Style Mixup        & 83.4 & 79.0 & 93.5  & 79.5 & 83.9  \\ 
\multicolumn{1}{c}{}               & Mixup       	     & 84.1 & 78.7 & 93.7  & 79.1 & 83.9  \\  
\multicolumn{1}{c}{}               & Ours    	         & \textbf{84.6} & \textbf{79.8} & \textbf{94.1}  & \textbf{80.1} & \textbf{84.7}   \\ 
\bottomrule
\end{tabular}
}
\end{center}
\vspace{-4mm}
\end{table*}

\vspace{-2mm}
\paragraph{Stopping criterion of attack}
Our algorithm stops the adversarial attack if the score of the ground-truth drops below a threshold, regardless of the number of attack iterations.
The threshold is a hyperparameter that balances the difficulty and plausibility of the synthesized style.
Table~\ref{tab:hyper}(b) presents the accuracies of our model by varying the threshold values.
Overall, our models with large thresholds achieve better results in general compared to the models with small ones. 
However, a small threshold is rather effective for the domains with large domain shifts, \eg, Clipart, because the augmented examples provide challenging styles.
Note that, if we set the threshold as too low, the overall performance is degraded substantially.
Similar to the ablation study about the number of attack iterations, the accuracies in the target domain and the source domain validation set have positive correlations when we vary the threshold $\tau$.
Hence, it is reasonable to select the proper hyperparameters based on the validation accuracy in the source domain.

\vspace{-2mm}
\paragraph{Step size}
Table~\ref{tab:eps} illustrates the results on Office-Home under the change of step size $\epsilon$.
As the $\epsilon$ increases, augmented styles get diverse while the plausibility is reduced.
Consequently, challenging target domains, \eg, Clipart, exhibit improved accuracies, while easy domains, \eg, Product and Real,  and source domains witness degraded performances.
Note that there is a positive correlation between the accuracies of source domain validation sets and target domains.

\vspace{-2mm}
\paragraph{Attack locations within models}
We evaluate the proposed algorithms by applying our style augmentation technique to multiple different layers in the network.
Table~\ref{tab:layer} demonstrates that the performance of the proposed method varies greatly depending on the location of style augmentation. 
Our algorithm provides a novel perspective on domain generalization in the sense that it attacks the feature statistics (styles) of the intermediate features instead of input images.
Table~\ref{tab:layer} clear shows inferior performance of ASA~\cite{zhong2022adversarial} and RGB-level style attacks in our method, highlighting the importance of feature-level augmentation.
Also, the comparison between RGB-level style attacks and CrossGrad~\cite{shankar2018generalizing} shows the importance of style attacks with style disentanglement.
Note that CrossGrad~\cite{shankar2018generalizing} does not disentangle styles but requires domain labels and domain classifiers.

\begin{figure}[t]
    \centering
            \includegraphics[width=0.85\linewidth]{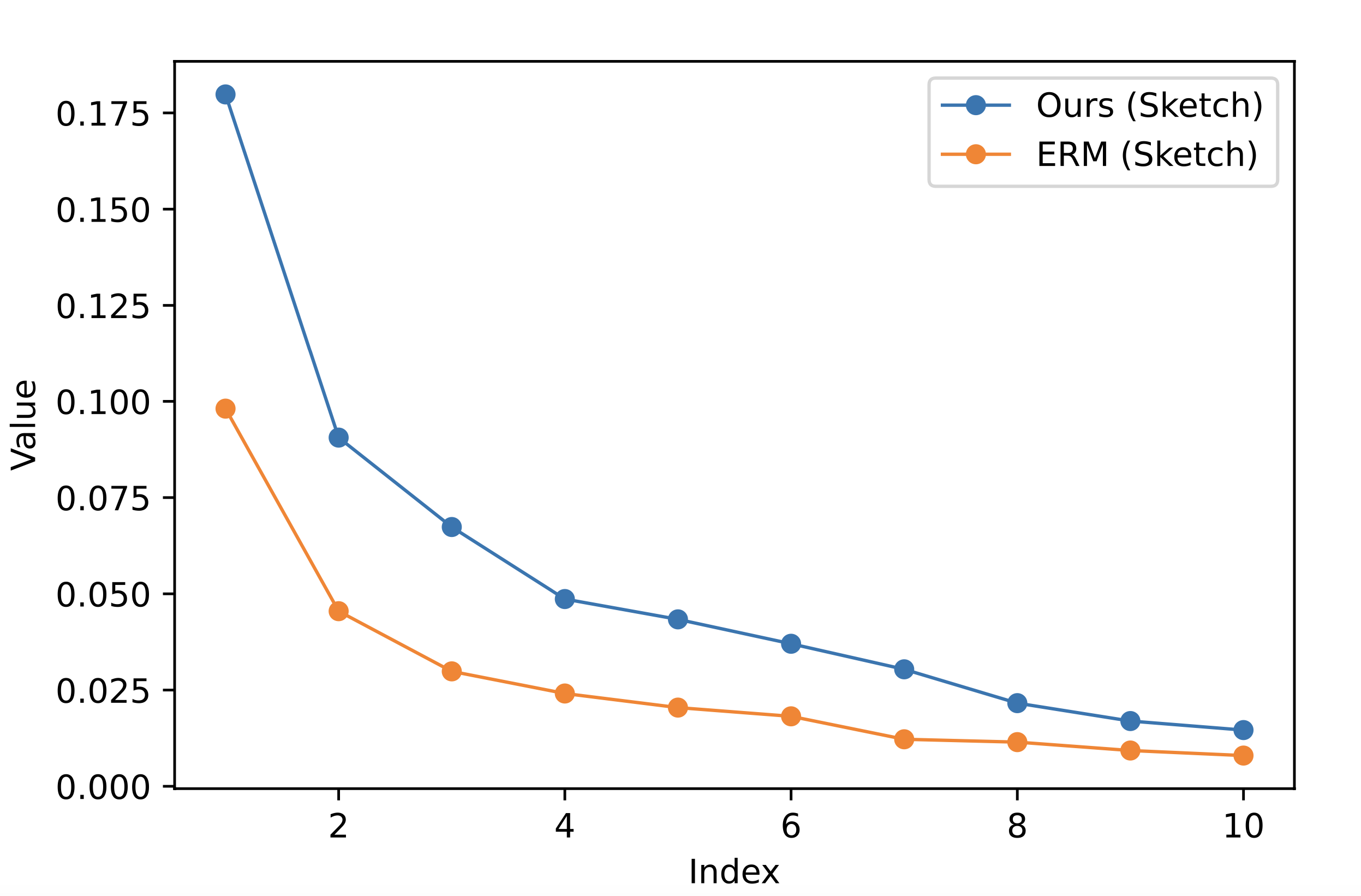}
\vspace{-2mm}
\caption{
    Eigenvalues of the covariance matrix of the channel mean vectors after the 2nd ResBlock for the test set of the target domain, sketch of the PACS dataset.
}
    \label{fig:eigen}
    \vspace{-2mm}
\end{figure}

\vspace{-2mm}
\paragraph{Attack objectives}
RASP randomly selects a target class for augmentation.
To show the effectiveness of this strategy, we test another option for augmentation directions.
One simple way is to add perturbation in the direction of decreasing the prediction score of the ground-truth, which is referred to as RASP$_\text{GT}$.
As shown in Table~\ref{tab:multi_abl}(a), such a direction is clearly outperformed by the proposed scheme.
We argue that this occurs because simply decreasing the score of the ground-truth label leads to a degenerate example, indicating an unrealistic image caused by untargeted style perturbations.
RASP$_\text{GT}$ produces more challenging but unrealistic styles.
Consequently, RASP$_\text{GT}$ improves accuracy on the Sketch domain, which is the most challenging domain while degrading performance on other domains.
This modified objective of the adversarial attack degrades the overall performance.

\vspace{-2mm}
\paragraph{Variations of NFM}
Table~\ref{tab:multi_abl}(b) presents the results from the various options related to the NFM modules.
When mixup is applied to styles or features instead of normalized features, classification accuracy significantly drops.
This is partly because the variations of our mixup strategy fail to benefit from the novel style generation capability of RASP.
When NFM is not employed, the accuracy of easy target domains, \eg, art and photo, is significantly degraded.
These results support our arguments that NFM maintains the knowledge learned from the source domains.

\vspace{-2mm}
\paragraph{Effects on diversity of styles}
We plot the eigenvalues of the covariance matrix of channel mean vectors after the 2nd ResBlock using the test set of the target domain, the sketch domain in PACS in this case. 
Figure~\ref{fig:eigen} illustrates that the proposed method with consistently larger eigenvalues allows the network to observe more diverse styles than the model based on ERM.

\section{Conclusion}
\label{sec:conclusion}	
We presented a simple yet effective style augmentation framework for domain generalization, called RASP, based on adversarial attacks.
RASP augments the styles that deceive the network by attacking the model itself.
Training models using the examples with the augmented styles is helpful for improving the generalization ability on unseen domains by making the feature extractors robust to the style changes.
In addition to this idea, we also proposed NFM to make the perturbed features have the desired properties and further enhance the domain generalization performance.
RASP with NFM does not require any architectural modifications or domain labels and can be easily incorporated into the existing baselines.
Extensive experiments show that the proposed algorithm consistently improves the generalization performance on unseen target domains across multiple datasets.

\paragraph{Acknowledgements}
This work was supported by Samsung Electronics Co., Ltd. [IO210917-08957-01], the National Research Foundation of Korea (NRF) grant [No. 2022R1A2C3012210] and Institute of Information \& communications Technology Planning \& Evaluation (IITP) grant [No.2022-0-00959, (Part 2) Few-Shot Learning of Causal Inference in Vision and Language for Decision Making; No.2021-0-01343, Artificial Intelligence Graduate School Program (Seoul National University)], funded by the Korea government (MSIT).

{\small
\bibliographystyle{ieee_fullname}
\bibliography{egbib}

\begin{thebibliography}{10}\itemsep=-1pt

\bibitem{vessl}
Jaeman An.
\newblock Model development with vessl, 2023.
\newblock Software available from vessl.ai.

\bibitem{balaji2018metareg}
Yogesh Balaji, Swami Sankaranarayanan, and Rama Chellappa.
\newblock Metareg: Towards domain generalization using meta-regularization.
\newblock In {\em NeurIPS}, 2018.

\bibitem{cha2021swad}
Junbum Cha, Sanghyuk Chun, Kyungjae Lee, Han-Cheol Cho, Seunghyun Park, Yunsung
  Lee, and Sungrae Park.
\newblock Swad: Domain generalization by seeking flat minima.
\newblock In {\em NeurIPS}, 2021.

\bibitem{chattopadhyay2020learning}
Prithvijit Chattopadhyay, Yogesh Balaji, and Judy Hoffman.
\newblock Learning to balance specificity and invariance for in and out of
  domain generalization.
\newblock In {\em ECCV}, 2020.

\bibitem{deng2009imagenet}
Jia Deng, Wei Dong, Richard Socher, Li-Jia Li, Kai Li, and Li Fei-Fei.
\newblock Imagenet: A large-scale hierarchical image database.
\newblock In {\em CVPR}, 2009.

\bibitem{du2020learning}
Yingjun Du, Jun Xu, Huan Xiong, Qiang Qiu, Xiantong Zhen, Cees~GM Snoek, and
  Ling Shao.
\newblock Learning to learn with variational information bottleneck for domain
  generalization.
\newblock In {\em ECCV}, 2020.

\bibitem{goodfellow2014explaining}
Ian~J Goodfellow, Jonathon Shlens, and Christian Szegedy.
\newblock Explaining and harnessing adversarial examples.
\newblock {\em ICLR}, 2015.

\bibitem{gulrajani2021search}
Ishaan Gulrajani and David Lopez-Paz.
\newblock In search of lost domain generalization.
\newblock In {\em ICLR}, 2021.

\bibitem{he2016deep}
Kaiming He, Xiangyu Zhang, Shaoqing Ren, and Jian Sun.
\newblock Deep residual learning for image recognition.
\newblock In {\em CVPR}, 2016.

\bibitem{huang2017arbitrary}
Xun Huang and Serge Belongie.
\newblock Arbitrary style transfer in real-time with adaptive instance
  normalization.
\newblock In {\em ICCV}, 2017.

\bibitem{izmailov2018averaging}
Pavel Izmailov, Dmitrii Podoprikhin, Timur Garipov, Dmitry Vetrov, and
  Andrew~Gordon Wilson.
\newblock Averaging weights leads to wider optima and better generalization.
\newblock In {\em UAI}, 2018.

\bibitem{kang2022style}
Juwon Kang, Sohyun Lee, Namyup Kim, and Suha Kwak.
\newblock Style neophile: Constantly seeking novel styles for domain
  generalization.
\newblock In {\em CVPR}, 2022.

\bibitem{kurakin2016adversarial}
Alexey Kurakin, Ian Goodfellow, and Samy Bengio.
\newblock Adversarial examples in the physical world.
\newblock {\em arXiv preprint arXiv:1607.02533}, 2016.

\bibitem{li2017deeper}
Da Li, Yongxin Yang, Yi-Zhe Song, and Timothy~M Hospedales.
\newblock Deeper, broader and artier domain generalization.
\newblock In {\em ICCV}, 2017.

\bibitem{li2018learning}
Da Li, Yongxin Yang, Yi-Zhe Song, and Timothy~M Hospedales.
\newblock Learning to generalize: Meta-learning for domain generalization.
\newblock In {\em AAAI}, 2018.

\bibitem{li2021simple}
Pan Li, Da Li, Wei Li, Shaogang Gong, Yanwei Fu, and Timothy~M Hospedales.
\newblock A simple feature augmentation for domain generalization.
\newblock In {\em ICCV}, 2021.

\bibitem{madry2018towards}
Aleksander Madry, Aleksandar Makelov, Ludwig Schmidt, Dimitris Tsipras, and
  Adrian Vladu.
\newblock Towards deep learning models resistant to adversarial attacks.
\newblock In {\em ICLR}, 2018.

\bibitem{moosavi2016deepfool}
Seyed-Mohsen Moosavi-Dezfooli, Alhussein Fawzi, and Pascal Frossard.
\newblock Deepfool: a simple and accurate method to fool deep neural networks.
\newblock In {\em CVPR}, 2016.

\bibitem{nuriel2021permuted}
Oren Nuriel, Sagie Benaim, and Lior Wolf.
\newblock Permuted adain: reducing the bias towards global statistics in image
  classification.
\newblock In {\em CVPR}, 2021.

\bibitem{peng2019moment}
Xingchao Peng, Qinxun Bai, Xide Xia, Zijun Huang, Kate Saenko, and Bo Wang.
\newblock Moment matching for multi-source domain adaptation.
\newblock In {\em ICCV}, 2019.

\bibitem{seo2020learning}
Seonguk Seo, Yumin Suh, Dongwan Kim, Geeho Kim, Jongwoo Han, and Bohyung Han.
\newblock Learning to optimize domain specific normalization for domain
  generalization.
\newblock In {\em ECCV}, 2020.

\bibitem{shankar2018generalizing}
Shiv Shankar, Vihari Piratla, Soumen Chakrabarti, Siddhartha Chaudhuri, Preethi
  Jyothi, and Sunita Sarawagi.
\newblock Generalizing across domains via cross-gradient training.
\newblock In {\em ICLR}, 2018.

\bibitem{tarvainen2017mean}
Antti Tarvainen and Harri Valpola.
\newblock Mean teachers are better role models: Weight-averaged consistency
  targets improve semi-supervised deep learning results.
\newblock {\em NeurIPS}, 2017.

\bibitem{ulyanov2016instance}
Dmitry Ulyanov, Andrea Vedaldi, and Victor Lempitsky.
\newblock Instance normalization: The missing ingredient for fast stylization.
\newblock {\em arXiv preprint arXiv:1607.08022}, 2016.

\bibitem{venkateswara2017deep}
Hemanth Venkateswara, Jose Eusebio, Shayok Chakraborty, and Sethuraman
  Panchanathan.
\newblock Deep hashing network for unsupervised domain adaptation.
\newblock In {\em CVPR}, 2017.

\bibitem{volpi2018generalizing}
Riccardo Volpi, Hongseok Namkoong, Ozan Sener, John~C Duchi, Vittorio Murino,
  and Silvio Savarese.
\newblock Generalizing to unseen domains via adversarial data augmentation.
\newblock In {\em NeurIPS}.

\bibitem{xie2019improving}
Cihang Xie, Zhishuai Zhang, Yuyin Zhou, Song Bai, Jianyu Wang, Zhou Ren, and
  Alan~L Yuille.
\newblock Improving transferability of adversarial examples with input
  diversity.
\newblock In {\em CVPR}, 2019.

\bibitem{xu2021fourier}
Qinwei Xu, Ruipeng Zhang, Ya Zhang, Yanfeng Wang, and Qi Tian.
\newblock A fourier-based framework for domain generalization.
\newblock In {\em CVPR}, 2021.

\bibitem{yang2021adversarial}
Fu-En Yang, Yuan-Chia Cheng, Zu-Yun Shiau, and Yu-Chiang~Frank Wang.
\newblock Adversarial teacher-student representation learning for domain
  generalization.
\newblock {\em NeurIPS}, 2021.

\bibitem{zhang2018mixup}
Hongyi Zhang, Moustapha Cisse, Yann~N Dauphin, and David Lopez-Paz.
\newblock mixup: Beyond empirical risk minimization.
\newblock In {\em ICLR}, 2018.

\bibitem{zhao2018generating}
Zhengli Zhao, Dheeru Dua, and Sameer Singh.
\newblock Generating natural adversarial examples.
\newblock In {\em ICLR}, 2018.

\bibitem{zhong2022adversarial}
Zhun Zhong, Yuyang Zhao, Gim~Hee Lee, and Nicu Sebe.
\newblock Adversarial style augmentation for domain generalized urban-scene
  segmentation.
\newblock In {\em NeurIPS}, 2022.

\bibitem{zhou2020deep}
Kaiyang Zhou, Yongxin Yang, Timothy Hospedales, and Tao Xiang.
\newblock Deep domain-adversarial image generation for domain generalisation.
\newblock In {\em AAAI}, 2020.

\bibitem{zhou2020learning}
Kaiyang Zhou, Yongxin Yang, Timothy Hospedales, and Tao Xiang.
\newblock Learning to generate novel domains for domain generalization.
\newblock In {\em ECCV}, 2020.

\bibitem{zhou2021domain}
Kaiyang Zhou, Yongxin Yang, Yu Qiao, and Tao Xiang.
\newblock Domain generalization with mixstyle.
\newblock In {\em ICLR}, 2021.

\end{thebibliography}
}

\end{document}